\documentclass[conference]{IEEEtran}
\IEEEoverridecommandlockouts
\usepackage{cite}
\usepackage{amsmath,amssymb,amsfonts}
\usepackage{algorithm}
\usepackage{graphicx}
\usepackage{algpseudocode}
\usepackage{textcomp}
\usepackage{xcolor}
\usepackage[hyphens]{url}

\def\BibTeX{{\rm B\kern-.05em{\sc i\kern-.025em b}\kern-.08em
T\kern-.1667em\lower.7ex\hbox{E}\kern-.125emX}}
\begin{document}

\title{A Stochastic Variance Reduced Nesterov's Accelerated Quasi-Newton Method \\
}

\author{
\IEEEauthorblockN{Sota Yasuda${}^\dag$, Shahrzad Mahboubi${}^\S$, S. Indrapriyadarsini${}^{\dag\dag}$, Hiroshi Ninomiya${}^\S$, Hideki Asai${}^{\dag\dag\dag}$}
\IEEEauthorblockA{${}^\dag$Faculty of Engineering, ${}^{\dag\dag}$ Graduate School of Science and Technology, ${}^{\dag\dag\dag}$ Research Institute of Electronics,\\Shizuoka University, Hamamatsu, Japan
\IEEEauthorblockA{${}^\S$Graduate School of Electrical and Information Engineering,\\ Shonan Institute of Technology, Fujisawa, Japan
\\\{yasuda.sota.16, s.indrapriyadarsini.17, asai.hideki\}@shizuoka.ac.jp}\{18T2012@sit, ninomiya@info\}.shonan-it.ac.jp}
}

\maketitle

\begin{abstract}
Recently algorithms incorporating second order curvature information have become popular in training neural networks. The Nesterov's Accelerated Quasi-Newton (NAQ) method has shown to effectively accelerate the BFGS quasi-Newton method by incorporating the momentum term and Nesterov's accelerated gradient vector. A stochastic version of NAQ method was proposed for training of large-scale problems. However, this method incurs high stochastic variance noise. This paper proposes a stochastic variance reduced Nesterov's Accelerated Quasi-Newton method in full (SVR-NAQ) and limited (SVR-LNAQ) memory forms. The performance of the proposed method is evaluated in Tensorflow on four benchmark problems - two regression and two classification problems respectively. The results show improved performance compared to conventional methods. 
\end{abstract}


\section{Introduction}
Neural networks have been effectively used in several applications to provide fast and accurate solutions. Most of these applications rely on large amounts of training data. Full batch strategies pose high computational cost and are hence not desirable for large scale problems. It is expected that the neural network utilizes lesser memory and computational load. Hence stochastic methods are more suitable for solving large scale optimization problems. This is because stochastic methods use a small subset of the dataset to give a probabilistic estimate and hence significantly reduce the per-iteration computational cost and storage memory. However, an inherent problem in stochastic methods is the presence of stochastic variance noise which slows down the convergence. Several works in machine learning and optimization have proposed stochastic variance reduced methods and have proven to effectively improve performance. Thus application of stochastic variance reduced methods in training neural networks could effectively improve the performance while maintaining low computational cost and memory.
\subsection{Related Works}
Gradient based algorithms are popular in training neural networks and can be categorized into first order and second or approximated second order methods. The gradient descent (GD) method is one of the earliest first order methods. However, GD uses full batch strategy and hence incurs high computational cost in large-scale problems. The stochastic gradient descent (SGD) method \cite{robbins1951stochastic} uses a small sub-sample of the dataset thus reducing the computational complexity. However, the convergence is much slower than GD due to stochastic variance \cite{wang2013variance}. 
Several stochastic variance reduction methods have been proposed in \cite{wang2013variance, shen2016adaptive,roux2012stochastic,shalev2013stochastic,johnson2013accelerating}. The stochastic variance reduced gradient (SVRG) method \cite{johnson2013accelerating} proposes a simple explicit variance reduction method. Furthermore, several first order methods such as AdaGrad \cite{duchi2011adaptive}, RMSprop \cite{tieleman2012lecture} and Adam \cite{kingma2014adam} have also shown to be effective in a stochastic setting. Among these, Adam is the most popular method and is shown to be highly stable. 

On the other hand, second order methods such as the Newton method incorporate curvature information and has shown to drastically improve convergence. However, computation of the inverse Hessian involves high computational cost and memory. Thus, several research works \cite{dennis1977quasi} focus on approximated second order methods that use an approximation of the curvature information and hence maintain low computational cost and memory. The BFGS quasi-Newton (QN) method \cite{nocedal2006} is a popular approximated second order method. Unlike in first order methods, it is challenging to get second order methods to work in a stochastic setting which is further augmented by stochastic variance noise. The o(L)BFGS method [13] is one of the earliest stable stochastic quasi-Newton methods, in which the gradients are computed twice using the same sub-sample to ensure stability and scalability. Several stochastic second order methods have also shown to have improved performance through variance reduction techniques such as in \cite{kolte2015accelerating, gower2016stochastic,gower2018stochastic,wang2017stochastic}. The SVRG+I and SVRG+II \cite{kolte2015accelerating} method was proposed as an acceleration of the SVRG method using second order curvature information. Specifically, the update of the curvature information of SVRG+I is followed by singular value thresholding, while in SVRG+II is updated by Broyden-Fletcher-Goldfarb-Shanon (BFGS).

The Nesterov's Accelerated quasi-Newton (NAQ) method \cite{ninomiya2017novel} is a recently proposed method that accelerates BFGS-QN by using the momentum term and Nesterov's accelerated gradient vector. Further a stochastic extension of the NAQ algorithm - o(L)NAQ \cite{oNAQ} was proposed and was shown to have improved performance compared to the o(L)BFGS method. This paper attempts to propose a stochastic variance reduced Nesterov's Accelerated quasi-Newton (SVR-NAQ) method. The proposed method is implemented on Tensorflow in its full and limited memory forms and the performance is compared against SVRG, Adam, o(L)NAQ and SVRG+II.
\section{Background}
Training in neural networks is an iterative process in which the parameters are updated in order to minimize an objective function. Given a mini-batch ${X \subseteq T_r}$ with samples ${(x_p,d_p)_{p \in X}}$ drawn at random from the training set ${T_r}$ and error function ${E_p({\bf w} ;x_p,d_p)}$ parameterized by a vector ${{\bf w} \in \mathbb{R}^d}$, the objective function is defined as 
\vspace{-1mm}
\begin{equation}\label{eq:obj} \underset{{\bf w} \in \mathbb{R}^d}{\text {min}} E({\bf w})= \frac {1}{b}{ \sum_{p \in X} E_p ({\bf w})}, 
\vspace{-1mm}
\end{equation}
where ${b}={|X|}$, is the batch size. In gradient based methods, the objective function ${E({\bf w})}$ under consideration is minimized by the iterative formula 
\vspace{-1mm}
\begin{equation}\label{eq:1} 
{\bf w}_{k+1} = {\bf w}_k + {\bf v}_{k+1}.
\vspace{-1mm}
\end{equation} 
where ${\it k}$ is the iteration count and ${\bf v}_{k+1}$ is the update vector, which is defined for each gradient algorithm. 
\subsection{Second Order Methods}
\subsubsection{BFGS quasi-Newton method}
Quasi-Newton (QN) methods utilize the gradient of the objective function to result in superlinear quadratic convergence. The BFGS algorithm is one of the most popular quasi-Newton methods for unconstrained optimization \cite{nocedal2006}. The update vector of the QN method is given as
\begin{equation}
{\bf v}_{k+1} = \alpha_k {\bf {g}}_k,
\end{equation} 
where the search direction ${\bf {g}}_k$ is given as
\begin{equation}
{\bf {g}}_k=-{\bf {H}}_k \nabla E({\bf w}_k).
\end{equation}

The Hessian matrix ${\bf H}_k$ is symmetric positive definite and is iteratively approximated by the BFGS formula \cite{nocedal2006},
\begin{equation}\label{eq:5}
{\bf H}_{k+1}= ( {\bf I}- {\bf s}_k {\bf y}_k^{\rm T}/{\bf y}_k^{\rm T} {\bf s}_k){\bf H}_k({\bf I}- {\bf y}_k {\bf s}_k^{\rm T}/{\bf y}_k^{\rm T} {\bf s}_k)+ {\bf s}_k {\bf s}_k^{\rm T}/{\bf y}_k^{\rm T} {\bf s}_k,
\end{equation}
where ${\bf I}$ denotes identity matrix, ${\bf s}_k = {\bf w}_{k+1} - {\bf w}_k$ and ${\bf y}_ k = \nabla E ( {\bf w}_{k+1} ) - \nabla E ({\bf w}_k)$. As the scale of the problem increases, the cost of computation and storage of the Hessian matrix becomes expensive. Limited memory scheme help reduce the cost considerably, especially in stochastic settings where the computations are based on small mini-batches of size ${b}$. In the limited memory LBFGS method, the Hessian matrix is defined by applying ${m}$ BFGS updates using only the last ${m}$ curvature pairs ${\{{\bf s}_k,{\bf y}_k\}}$, where $m$ denotes the memory size. The search direction ${\bf g}_k$ is evaluated using the two-loop recursion \cite{nocedal2006}.

\subsubsection{ SVRG+II}
 	 Acceleration of SVRG method using second order information was proposed in \cite{kolte2015accelerating}. While SVRG+I used sub-sampled Hessian, SVRG+II uses LBFGS for the Hessian approximation. Since the focus of this paper revolves around BFGS based method, the SVRG+II method is briefly explained below.

    The SVRG+II method combines SVRG and (L)BFGS method. The algorithm is shown in Algorithm 1. The curvature pair information $\{{\bf s}_k, {\bf y}_k\}$ is computed only once per epoch. The curvature information pair $\{{\bf s}_k, {\bf y}_k\}$ is expressed as .
\begin{equation}
{\bf s}_k = {\bf w}_{k+1}  - {\bf w}_k  
\end{equation}
\begin{equation}
{\bf y}_k={\bf \Omega}_{k+1} -{\bf \Omega}_k 
\end{equation}
where ${\bf \Omega}_k$ is the full gradient computed over the entire dataset and $k$ represents the epoch.
The search direction term in each iteration given by ${\bf {g}}_t = -{\bf H}_k {\bf f}_t $  is updated using the two-loop recursion. Computation of ${\bf f}_t $ involves both the mini-batch gradient ${\bf \nabla E}_{i_t } ({\bf x}_t )$ and the full gradient ${\bf \Omega}_k$ 
The update vector of SVRG+II is given by:
\begin{equation}
{\bf g}_{t}= -{\bf H}_k ({\bf \nabla E}_{i_t} ({\bf x}_t)-{\bf \nabla E}_{i_t } ({\bf w}_{k+1} )+{\bf \Omega}_{k+1} )
\end{equation}
where $t$ represents the iteration count within an epoch and ${\bf \nabla E}_{i_t } ({\bf w} )$ denotes the mini-batch gradient. Note that the Hessian matrix is thus fixed throughout an epoch. 
The first epoch of SVRG+II is run as an epoch of SVRG to obtain the $\{{\bf s}_k,{\bf y}_k\}$ curvature pair information. \\\vspace{-4mm}
\begin{algorithm}[htb]
\centering
\caption{SVRG+II}
\begin{algorithmic}[1]
\label{Algo:SVRg-II}
\Ensure ${\bf w_0}$, $k = 0$ 
\Require Run SVRG for the first epoch to get ${\bf w}_1$
\While {$k<k_{max}$}
\State Compute full gradient ${\bf \Omega}_{k+1} \leftarrow \nabla E({\bf w}_{k+1})$
\State ${\bf s}_k = {\bf w}_{k+1} - {\bf w}_k$
\State ${\bf y}_k = {\bf \Omega}_{k+1} - {\bf \Omega}_{k}$
\State ${\bf x}_0 \leftarrow {\bf w}_{k+1}$
\State Compute ${\bf H}_{k+1} $ using (5)
\For {$t=0,1,2,...,n$}
\State Sample $i_t$ from [1:n] uniformly random
\State ${\bf f}_t \leftarrow \nabla E_{i_t} ({\bf x}_t ) - \nabla E_{i_t}({\bf w}_{k+1}) + {\bf \Omega}_{k+1}$
\State ${\bf g}_t =-{\bf H}_{k+1}{\bf f}_t$
\State ${\bf x}_{t+1} \leftarrow {\bf x}_t + \alpha_t {\bf g}_t$
\EndFor
\State $ k = k + 1$ 
\State ${\bf w}_{k+1} \leftarrow {\bf x}_n$
\EndWhile
\end{algorithmic}
\end{algorithm}
\vspace{-3mm}

\section{Proposed Algorithm - Stochastic Variance Reduced Nesterov's Accelerated quasi-Newton (SVR-(L)NAQ)}

The NAQ \cite{ninomiya2017novel} method achieves faster convergence compared to the standard QN by quadratic approximation of the objective function at ${\bf w}_k+\mu {\bf v}_k$ and by incorporating the Nesterov's accelerated gradient $\nabla E({\bf w}_k+\mu {\bf v}_k)$ in its Hessian update. The update vector of NAQ is given as 
\begin{equation}
{\bf v}_{k+1} = \mu_k {\bf v}_k + \alpha_k {\bf {g}}_k,
\end{equation}
\begin{equation}
{\bf {g}}_k=-{\bf {H}}_k \nabla E({\bf w}_k+\mu_k {\bf v}_k).
\end{equation}
The Hessian matrix ${\bf H}_{k+1}$ in NAQ is symmetric, positive definite matrix and is iteratively approximated by 
\begin{equation}
{\bf H}_{k+1}= ( {\bf I}- {\bf p}_k {\bf q}_k^{\rm T}/{\bf q}_k^{\rm T} {\bf p}_k){\bf H}_k({\bf I}- {\bf q}_k {\bf p}_k^{\rm T}/{\bf q}_k^{\rm T}{\bf p}_k)+ {\bf p}_k {\bf p}_k^{\rm T}/{\bf q}_k^{\rm T} {\bf p}_k,
\end{equation}
where ${\bf p}_k = {\bf w}_{k+1} - ({\bf w}_k+ \mu{\bf v}_k)$ and ${\bf q}_k = \nabla E ( {\bf w}_{k+1} ) - \nabla E ({\bf w}_k+ \mu {\bf v}_k)$. It can be observed that NAQ involves two gradient calculations per iteration in ${\bf q}_k$, which adds to the computational cost compared to the BFGS method. However, this is well compensated due to the acceleration using Nesterov's Accelerated gradient.\\ 
\begin{algorithm}[ht]
\centering
\caption{Proposed method : SVR-NAQ }
\begin{algorithmic}[1]
\label{Algo:SVRNAQ}
\Ensure ${\bf w}$ $ \leftarrow $ uniform random $\{-0.5,0.5\}$, ${\bf V}_0=0$, $k = 0$
\Require Run SVRG for the first epoch to get ${\bf w}_0$
\While {$k<k_{max}$}
\State Compute full gradient ${\bf \Omega}_{k+1} \leftarrow \nabla E({\bf w}_{k+1})$
\State ${\bf p}_k = {\bf w}_{k+ 1}- ({\bf w}_k + \mu {\bf V}_k)$
\State ${\bf q}_k = {\bf \Omega}_{k+1} - \nabla E({\bf w}_k+\mu {\bf V}_k)$
\State ${\bf x}_0 \leftarrow {\bf w}_{k+1}$
\State ${\bf v}_0 \leftarrow {\bf V}_k$
\State Compute ${\bf H}_{k+1} $ using (11)
\For {$t=0,1,2,...,n$}
\State Sample $i_t$ from [1:n] uniformly random
\State ${\bf f}_t \leftarrow \nabla E_{i_t} ({\bf x}_t + \mu {\bf v}_t) - \nabla E_{i_t}({\bf w}_{k+1}) + {\bf \Omega}_{k+1}$
\State ${\bf g}_t = -{\bf H}_{k+1} {\bf f}_t$
\State ${\bf v}_{t+1} \leftarrow \mu {\bf v}_t + \alpha_t {\bf g}_t$
\State ${\bf x}_{t+1} \leftarrow {\bf x}_{t} + {\bf v}_{t+1}$
\EndFor
\State ${\bf V}_{k+1} \leftarrow {\bf v}_n$
\State $ k = k + 1$ 
\State ${\bf w}_{k+1} \leftarrow {\bf x}_n$
\EndWhile
\end{algorithmic}
\end{algorithm}
\indent
In this paper, a Stochastic Variance Reduced Nesterov's Accelerated quasi-Newton (SVR-NAQ) method is proposed to accelerate and improve the performance of training for large-scale problems. The update vector of the proposed algorithm is the same as (11). However, ${\bf g}_t$ is given as 
\begin{equation}
{\bf {g}}_t=-{\bf {H}}_{k+1} \nabla E_{i_t} ({\bf x}_{t} + \mu {\bf v}_{t}) - \nabla E_{i_t}({\bf w}_{k+1}) + {\bf \Omega}_{k+1},
\end{equation}
where $k, t$ and $\Omega_k= \nabla E({\bf w}_k)$ denotes the epoch, iteration count and full gradient respectively. The Hessian matrix ${\bf \rm H}_k$ in SVR-NAQ is iteratively approximated by (11). Note that the curvature pair information $\{{\bf p}_k, {\bf q}_k\}$, required for the Hessian update is computed once per epoch and ${\bf q}_k$ uses the full gradient. 
\begin{equation}
{\bf q}_k= {\bf \Omega}_{k+1} - \nabla E({\bf w}_k+\mu {\bf V}_k)
\end{equation}
Thus the proposed SVR-NAQ is more effective compared to oNAQ as the Hessian matrix is updated with less noise. 
To compute the $\{{\bf p}_k, {\bf q}_k\}$ curvature pair information in the first epoch, SVR-NAQ is pre-trained by SVRG to obtain ${\bf w}_1$.
A polynomial decay schedule is used to obtain the step size $\alpha_t$ which is given by 
\begin{equation}
\alpha_t= {\alpha_0}/{\sqrt{t}}.
\end{equation}
where $\alpha_0$ is usually set to 1. The algorithm of proposed SVR--NAQ is shown in Algorithm 2. Further, the limited memory version of SVR-NAQ (SVR-LNAQ) can be realized by updating the search direction using the two-loop recursion i.e. ${\bf g}_t$ in step 11 of Algorithm 2 is determined by Algorithm 3 using ${\bf f}_t$ and $\{P,Q\}$. Thus step 7 of Algorithm 2 can be omitted and instead the last ${\it m}$ curvature pair $\{{\bf p}_k, {\bf q}_k\}$ is stored in $\{P,Q\}$ buffer of size ${\it m}$. To improve the performance by average sampling noise, ${{\bf H}^{(0)}_k}$ in step 6 of Algorithm 3 is computed using (14).
\begin{equation}
\label{eq:oLBFGS}
{{\bf H}^{(0)}_k} = \displaystyle\frac{1}{\rm min(k,m)}\displaystyle\sum^{\rm min(t,m)}_{i=1} \frac{{\bf p}_{k-i}^{\rm T} {\bf q}_{k-i}}{{\bf q}_{k-i}^{\rm T} {\bf q}_{k-i}}
\end{equation}
\begin{algorithm}[t]
\caption{ Direction Update Two-loop Recursion}
\begin{algorithmic}[1]
\label{Algo:dirUp}
\Require gradient vector ${\bf f}_t$, curvature pair $\{P,Q\}$ buffer
\State $\tau = {\rm length}(P)$
\For {$i={\tau},\ldots,2,1~~$}
\State${\sigma}_i=({{\bf p}_i^{\rm T} {\bf {\eta}}_t})/({{\bf q}_i^{\rm T} {\bf p}_i})$
\State ${\bf {g}}_t={\bf {f}}_t-{ \sigma}_i {\bf q}_{i}$
\EndFor
\State ${\bf {g}}_t= {{\bf H}^{(0)}_k}{\bf {g}}_t$
\For {$i:=1, 2, ..., {\tau~~}$}
\State${\beta}=({{\bf q}_i^{\rm T} {\bf {g}}_t})/({{\bf q}_i^{\rm T} {\bf p}_i})$
\State${\bf {g}}_t={\bf {g}}_t-({\sigma}_i-{\beta}){\bf p}_i$
\EndFor
\State ${\bf \eta}_t = -{\bf {g}}_t$
\State{\bf return}\;${\bf {\eta}}_t$
\end{algorithmic}
\end{algorithm}

\section{Simulation Results}
The performance of the proposed SVR-(L)NAQ methods are evaluated on four benchmark problems - two regression and two classification problems. For the regression problem, the white wine quality\cite{cortez2009modeling} and CASP \cite{rana2013physicochemical} dataset are chosen and for the classification problem, the MNIST \cite{mnist2010data} and CIFAR-10 \cite{krizhevsky2009learning} datasets are chosen. The regression problems are evaluated on Multi-Layer Neural Network (MLNN) and classification are evaluated on a simple Convolution Neural Network (CNN). The details of the simulation are given in Table \ref{tab:style}. The performance of the proposed algorithm is compared against Adam \cite{kingma2014adam} , SVRG \cite{shen2016adaptive} , SVRG-II \cite{kolte2015accelerating} and o(L)NAQ \cite{oNAQ}. All the algorithms are implemented on Tensorflow using the ScipyOptimizerInterface class. The weights are initialized in the range [-0.5,0.5] with uniform random distribution. The hyperparameters for Adam, SVRG, SVRG-II and o(L)NAQ are set to the values suggested in the respective of each paper, respectively. The step size $\alpha$ is chosen to be 0.025 for SVRG and the first epoch of SVR-(L)NAQ. The performance of o(L)NAQ and SVR-(L)NAQ for different values of the momentum coefficient $\mu_k$ were studied and the results using best value of $\mu_k$ is presented.
\begin{table*}[htb]
\vspace{-0.3cm}
\begin{center}
\caption{Details of the Simulation}
\label{tab:style}
\fontsize{9}{11}\selectfont
\begin{tabular}{ccccccc}
\hline
\\[-3mm]
{\;\;\; \;\;\;} & {\;\;\; Wine Quality \;\;\;} & {\;\;\; CASP \;\;\;}& {\;\;\;\; MNIST \;\;\;\;} & {\;\;\;\; CIFAR-10 \;\;\;\;}\\
\hline
\\[-3mm]
{task } & regression & regression & classification & classification \\
{input } & 11 & 9 & 28x28x1 & 32x32x3\\
{NN structure} & 11-10-4-1 & 9-10-6-1 & $C_{3,3}-C_{3,5}-FC_{10}$ & $C_{5,3}-C_{5,5}-FC_{16}-FC_{32}$ \\
{parameters (\it d)} & 169 & 173 & 6,028 & 13,170 \\
{train set} & 3,918 & 36,584 & 55,000 & 50,000 \\
{test set} & 980 & 9,146 & 10,000 & 10,000\\
{classes/output } & 1 & 1 & 10 & 10 \\
{batch size ($b$)} & 8/32 & 16/64 & 64 & 128\\
{memory ($m$)} & 4 & 4 & 4 & 4\\
\hline
\end{tabular}
\end{center}
\vspace{-4mm}
\end{table*}

\begin{figure}[t]
\vspace{-5mm}
\begin{center}
\includegraphics[width=8cm]{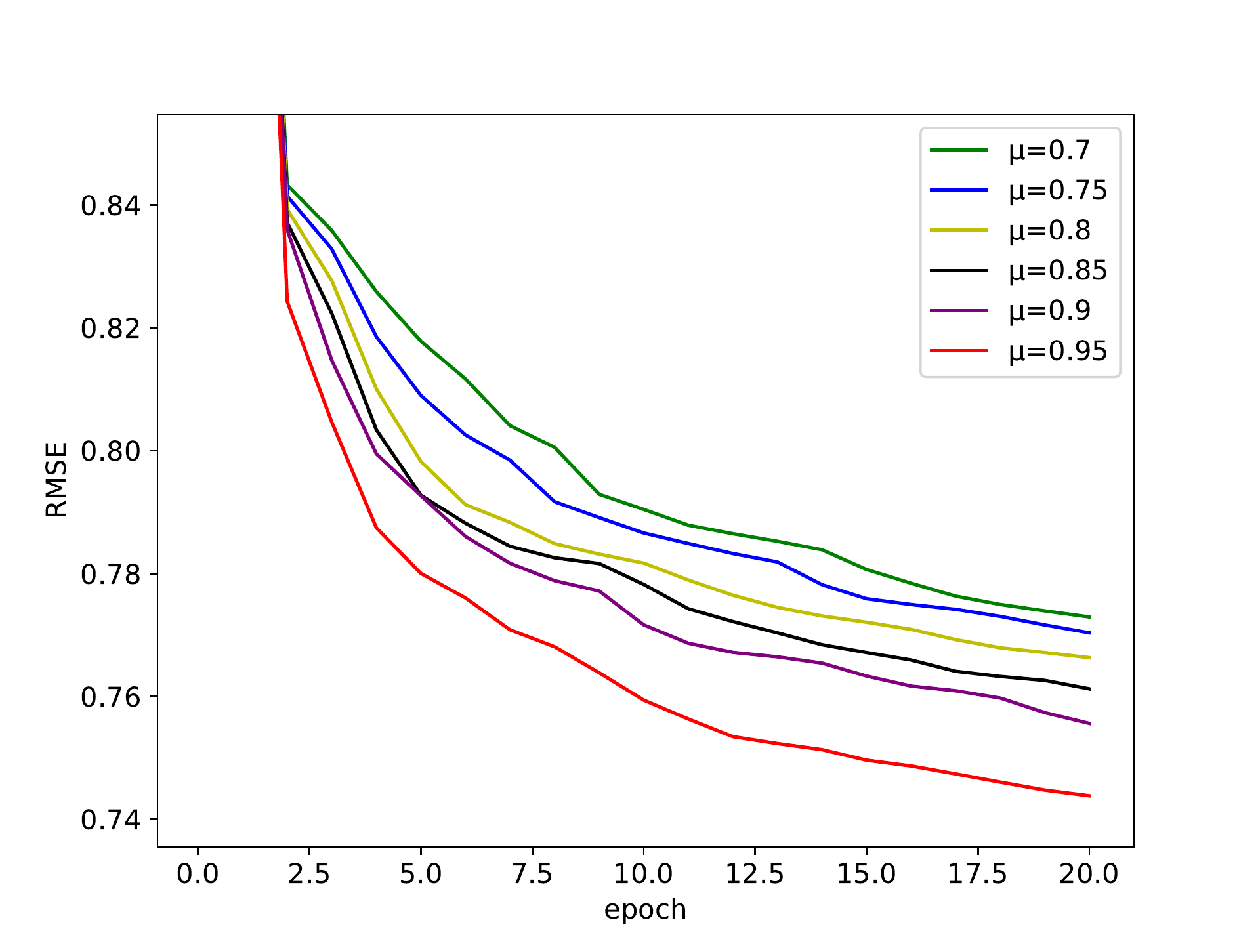}
\end{center}
\vspace{-5mm}
\caption{Effect of the parameter $\mu_k$ in white wine quality dataset}
\label{fig:mu_of_SVR-NAQ}
\vspace{-5mm}
\end{figure}

\subsection{Results on Regression}
The performance of the proposed methods was evaluated on white wine quality \cite{cortez2009modeling} and CASP \cite{rana2013physicochemical}
datasets using a 3-layered neural network. The hidden layers use sigmoid activation function while the output layer use linear activation function. Mean squared error function was used. Both datasets were z-normalized to have zero mean and unit variance. The datasets are split in 80-20\% for train and test set, respectively. The maximum epochs was set to 20. To validate the efficiency of the proposed variance reduced method, the performance of SVR-(L)NAQ is compared with Adam, SVRG, SVRG+II and o(L)NAQ with batch sizes proposed in \cite{oNAQ} and smaller batch size.

\begin{figure}[b]
\vspace{-9mm}
\begin{center}
\includegraphics[width=8cm]{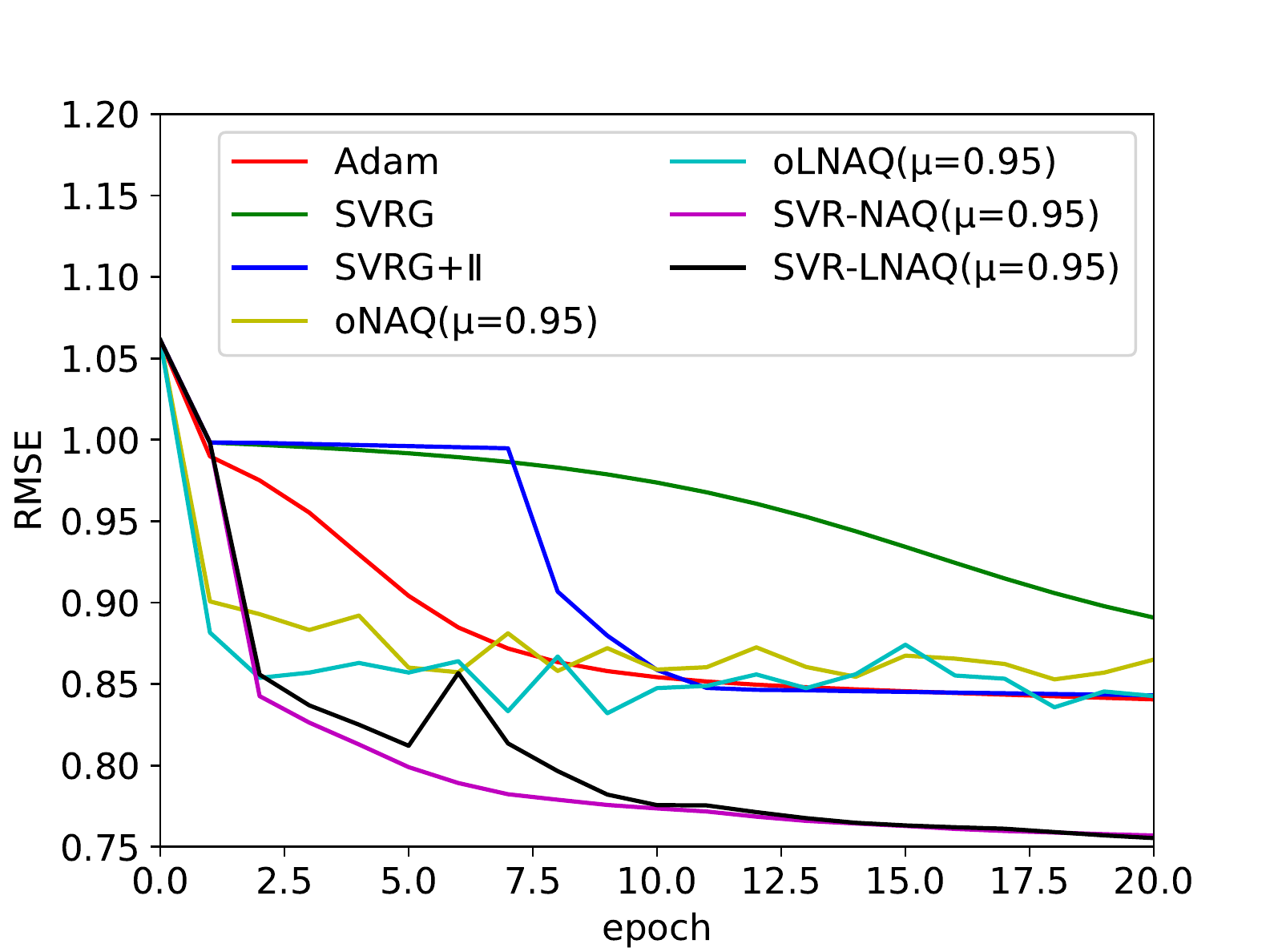}
\end{center}
\vspace{-3mm}
\caption{Results of white wine quality dataset for $b = 32$.}
\label{fig:wine_32_train_loss.pdf}
\vspace{-6mm}
\end{figure}

\subsubsection{White Wine Quality dataset}
The task in the white wine quality problem \cite{cortez2009modeling} is to estimate the quality of white wine on a scale of 3 to 9 based on 11 physiochemical test values. The momentum coefficient for o(L)NAQ \cite{oNAQ} and SVR-(L)NAQ is set to $\mu_k = 0.95$. Fig. \ref{fig:mu_of_SVR-NAQ} shows the performance of SVR-NAQ for different values of the momentum coefficient. It can be observed that $\mu_k = 0.95$ converges with a small RMSE much faster than other values. By similar observations conducted, the parameters for the other problems are also chosen.

\begin{figure}[b]
\vspace{-9mm}
\begin{center}
\includegraphics[width=8cm]{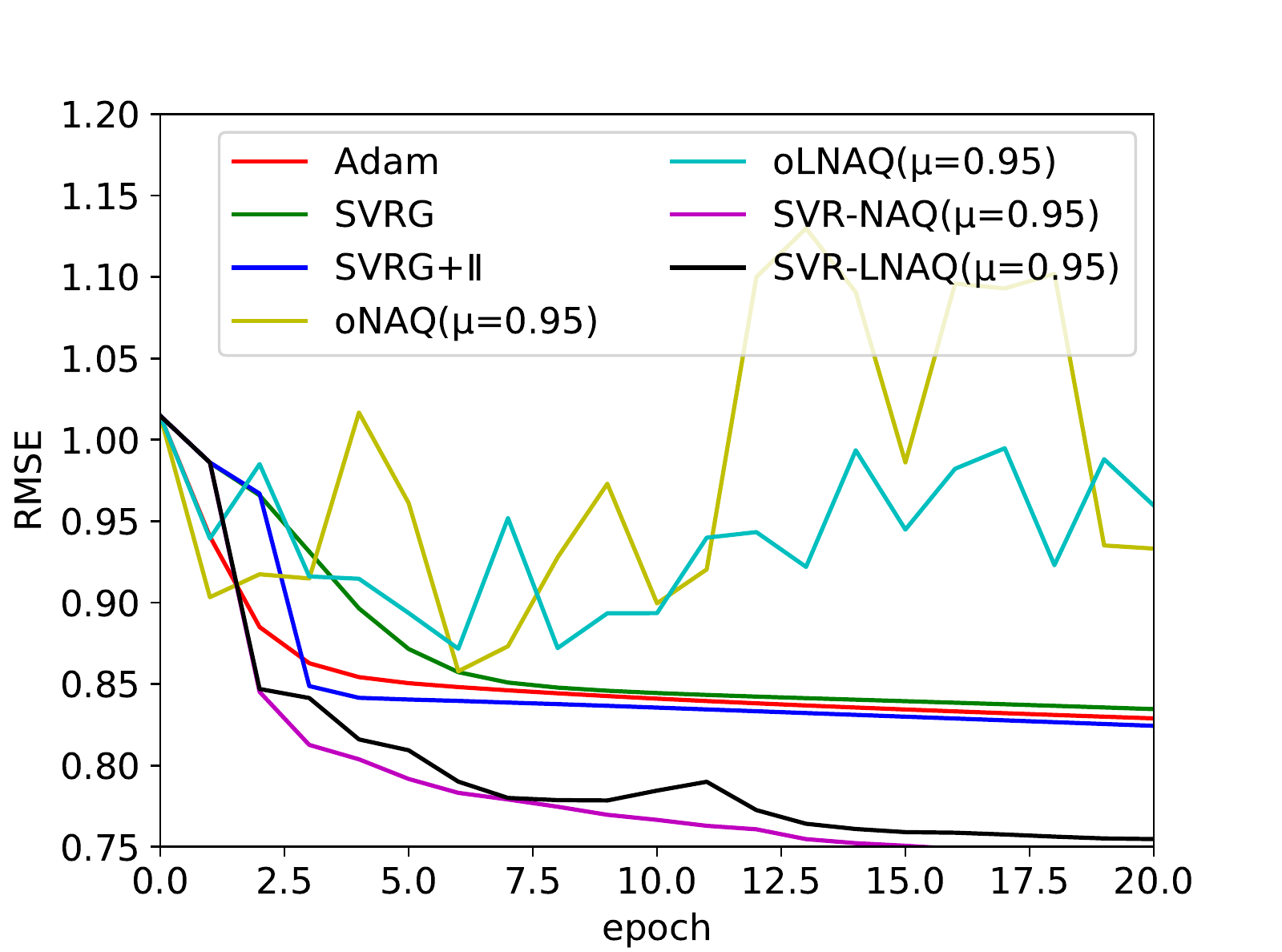}
\end{center}
\vspace{-3mm}
\caption{Results of white wine quality dataset for $b = 8$.}
\label{fig:wine_8_train_loss.pdf}
\vspace{-6mm}
\end{figure}

The results of RMSE vs epochs for white wine quality problem for batch sizes 32 and 8 are shown in Fig. \ref{fig:wine_32_train_loss.pdf} and \ref{fig:wine_8_train_loss.pdf} respectively. For $b=32$ , it can be observed that the o(L)NAQ RMSE reduces at a faster pace in the initial epoch and gradually remains constant. On the other hand, the proposed SVR-(L)NAQ reduces to much smaller errors fast but after the initial epoch. This is because the first epoch is evaluated by SVRG. For a smaller batch size of $b = 8$, it can be observed that o(L)NAQ does not perform well for smaller batch sizes while the variance reduction schemes perform well. The proposed SVR-(L)NAQ method can attain much smaller errors even for small batch sizes, thus confirming its efficiency over o(L)NAQ and the other variance reduced methods in consideration.

\begin{figure*}[ht]
\begin{minipage}[t]{0.49\textwidth}
\includegraphics[width=8cm]{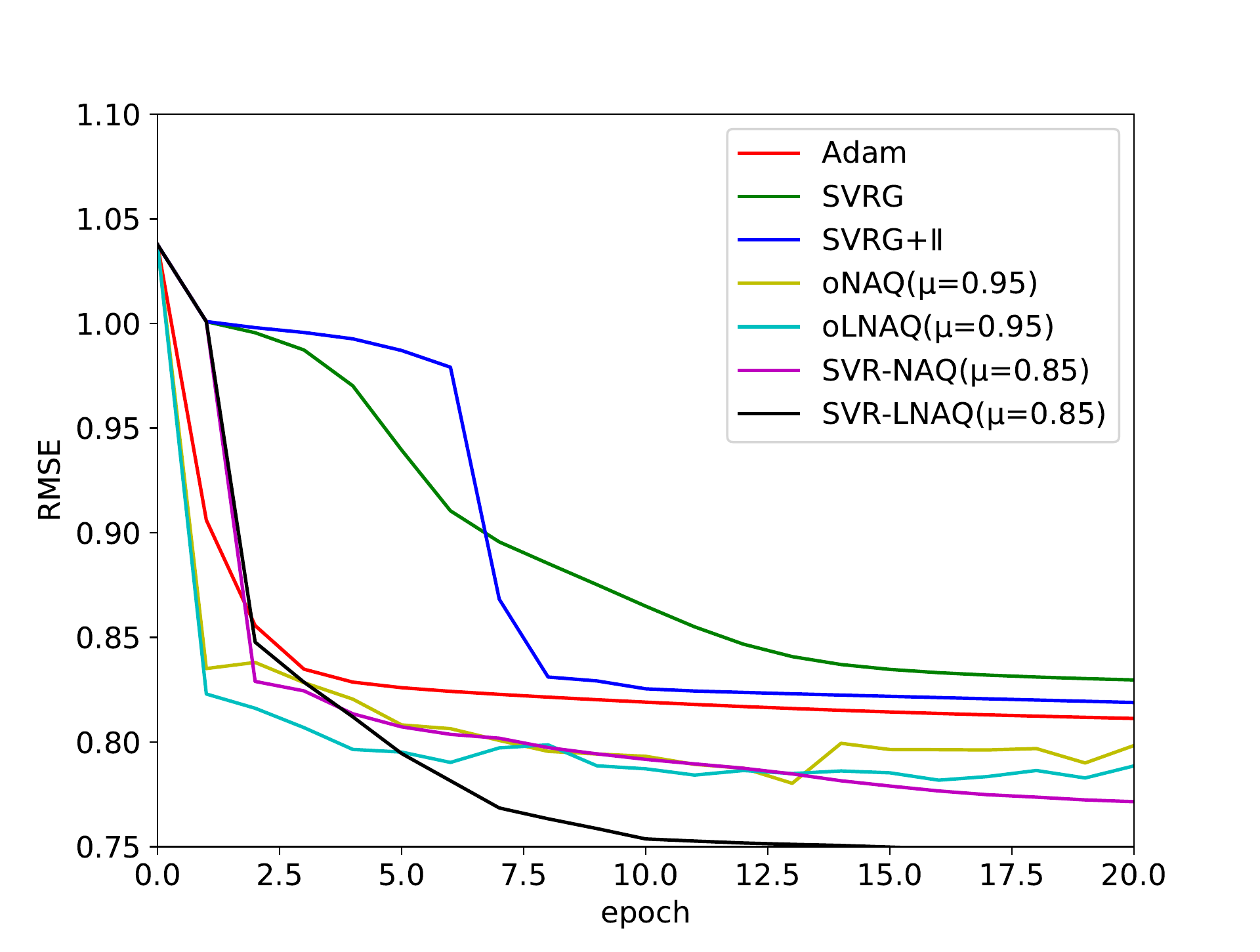}
\vspace{-3mm}
\caption{Results of CASP detaset for $b = 64$.}
\label{fig:CASP64}
\end{minipage}
\begin{minipage}[t]{0.49\textwidth}
\includegraphics[width=8cm]{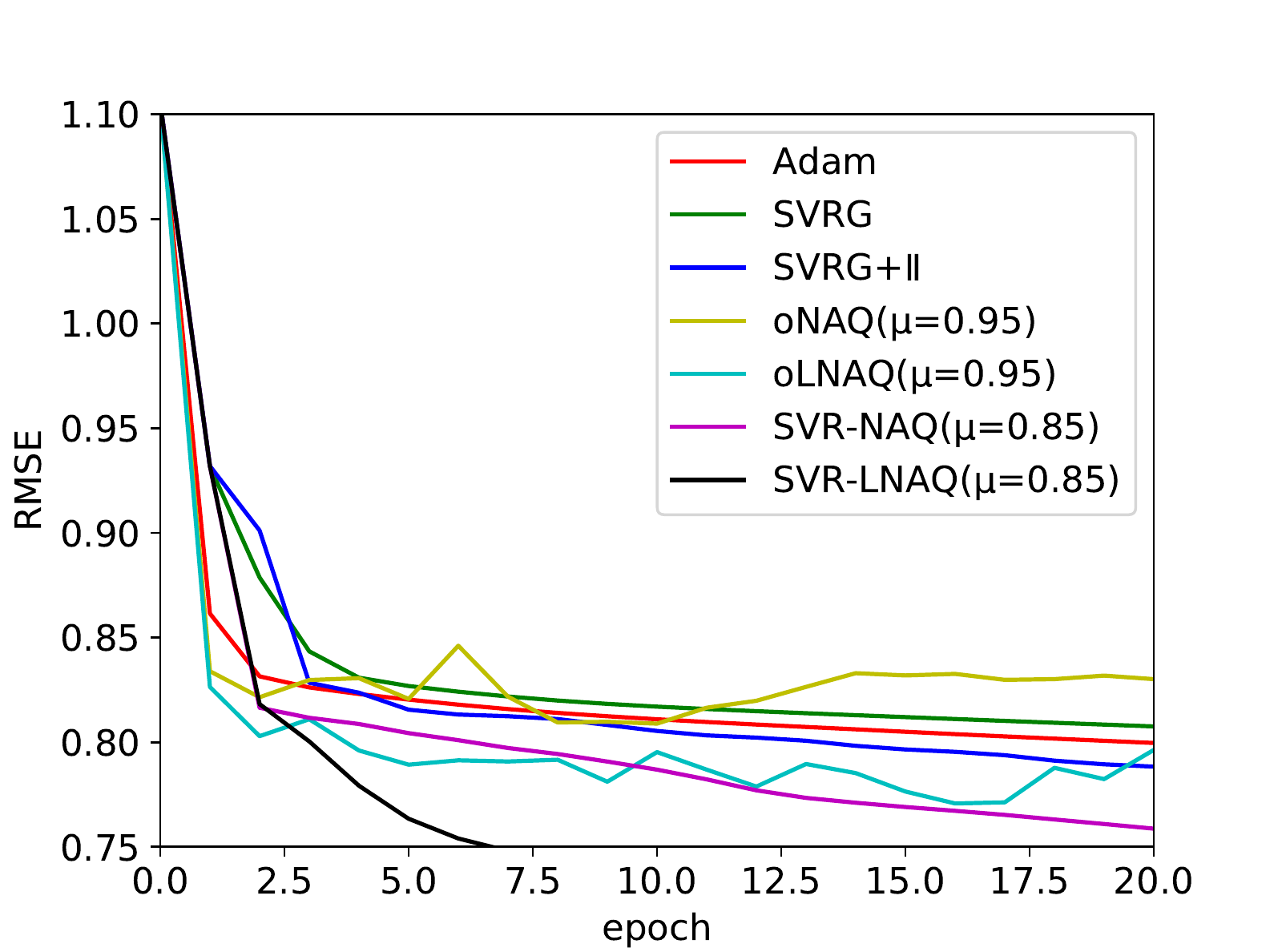}
\vspace{-2mm}
\caption{Results of CASP detaset for $b = 16$.}
\label{fig:CASP16}
\end{minipage}
\vspace{-4mm}
\end{figure*}
\subsubsection{Results on CASP dataset}
The performance of the proposed method is further evaluated on a bigger problem such as the CASP (Critical Assessment of protein Structure Prediction) dataset \cite{rana2013physicochemical} where the task is to predict the protein structure based on the physiochemical properties of protein tertiary structure. 
The results of RMSE vs epochs for batch sizes 64 and 16 are shown in Fig. \ref{fig:CASP64} and \ref{fig:CASP16} respectively. 
The $\mu$ for o(L)NAQ and SVR-(L)NAQ were set as $\mu=0.95$ and $\mu=0.85$, respectively. From the results, it can be observed that o(L)NAQ and SVR-(L)NAQ methods can attain much lower error compared to Adam, SVRG and SVRG+II, thus confirming the effectiveness of the momentum acceleration. Similar to the results in the previous example, o(L)NAQ is faster is the initial epochs but does not reduce to smaller errors on further increasing the epochs. However, the proposed SVR-(L)NAQ method attain much lower errors as the epoch progresses. On the other hand, the other algorithms fail to reduce to smaller errors even upon increasing the epoch.  

In order to evaluate the effectiveness of the proposed algorithm on a larger problem, the next section deals with classification problems. 

\begin{figure}[b]
\vspace{-8mm}
\begin{center}
\includegraphics[width=8cm]{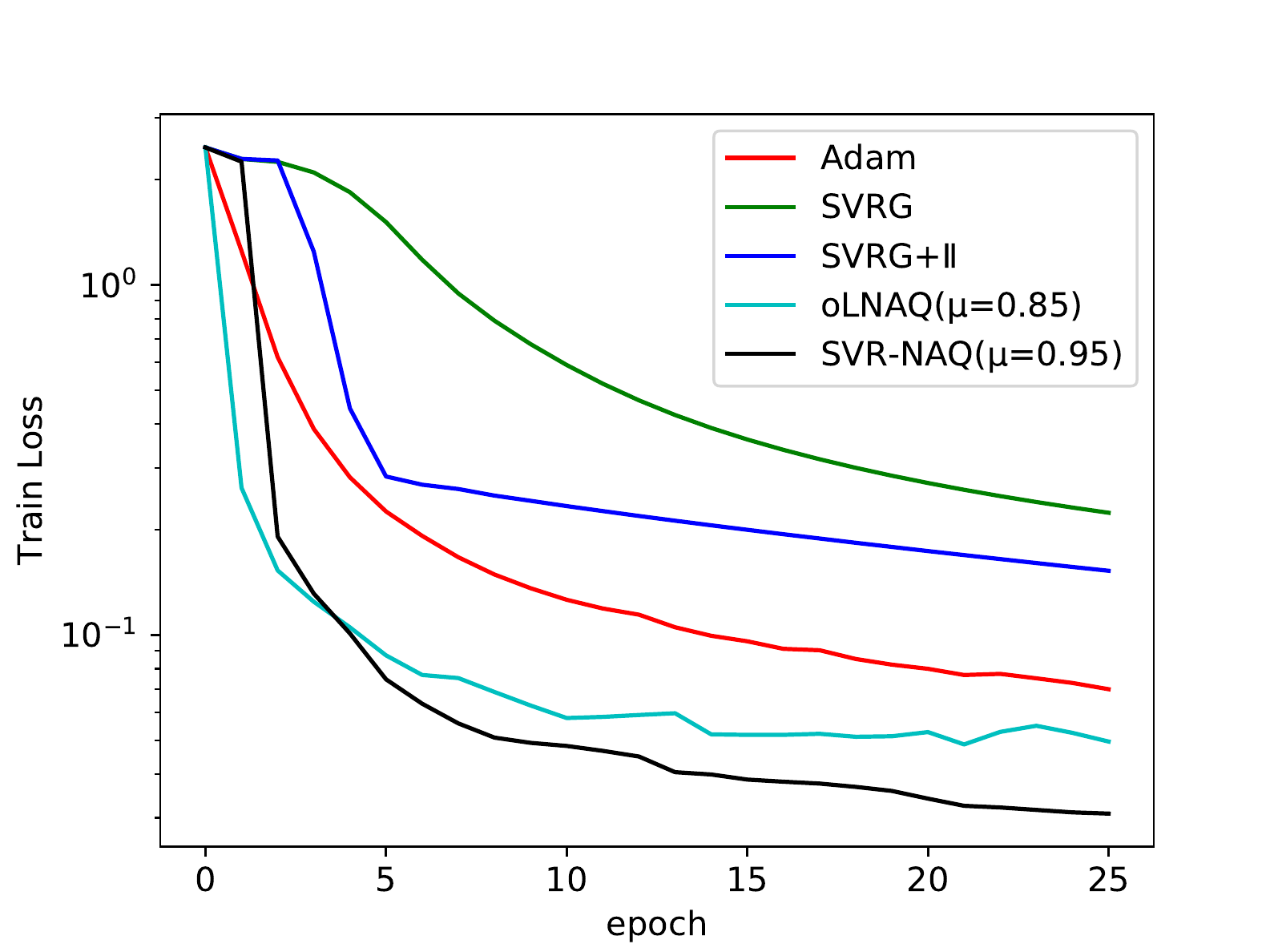}
\end{center}
\vspace{-5mm}
\caption{Train loss of MNIST for $b = 64$}
\label{fig:mnistTRL}
\vspace{-6mm}
\end{figure}
\subsection{Results on Classification}
Further, the performance of the proposed method on classification problems is evaluated. For the classification task, standard image classification MNIST and CIFAR-10 dataset is chosen. The performance is evaluated on a simple convolution neural network (CNN). The CNN architecture chosen comprises of two convolution layers, each followed by a 2x2 max pooling layer with a stride of 2. The CNN structure is illustrated in Table \ref{tab:style}. The convolution filter kernel and number of channels are represented as $C_{k,c}$ and fully connected layers are represented as $FC_{h}$ where $h$ is the number of hidden neurons. Sigmoid activation function and softmax cross entropy error function is used. Owing to the large number of parameters ($d$) and system constraints, the performance of the algorithms in only its limited memory version are illustrated.   
\begin{figure}[b]
\vspace{-8mm}
\begin{center}
\includegraphics[width=8cm]{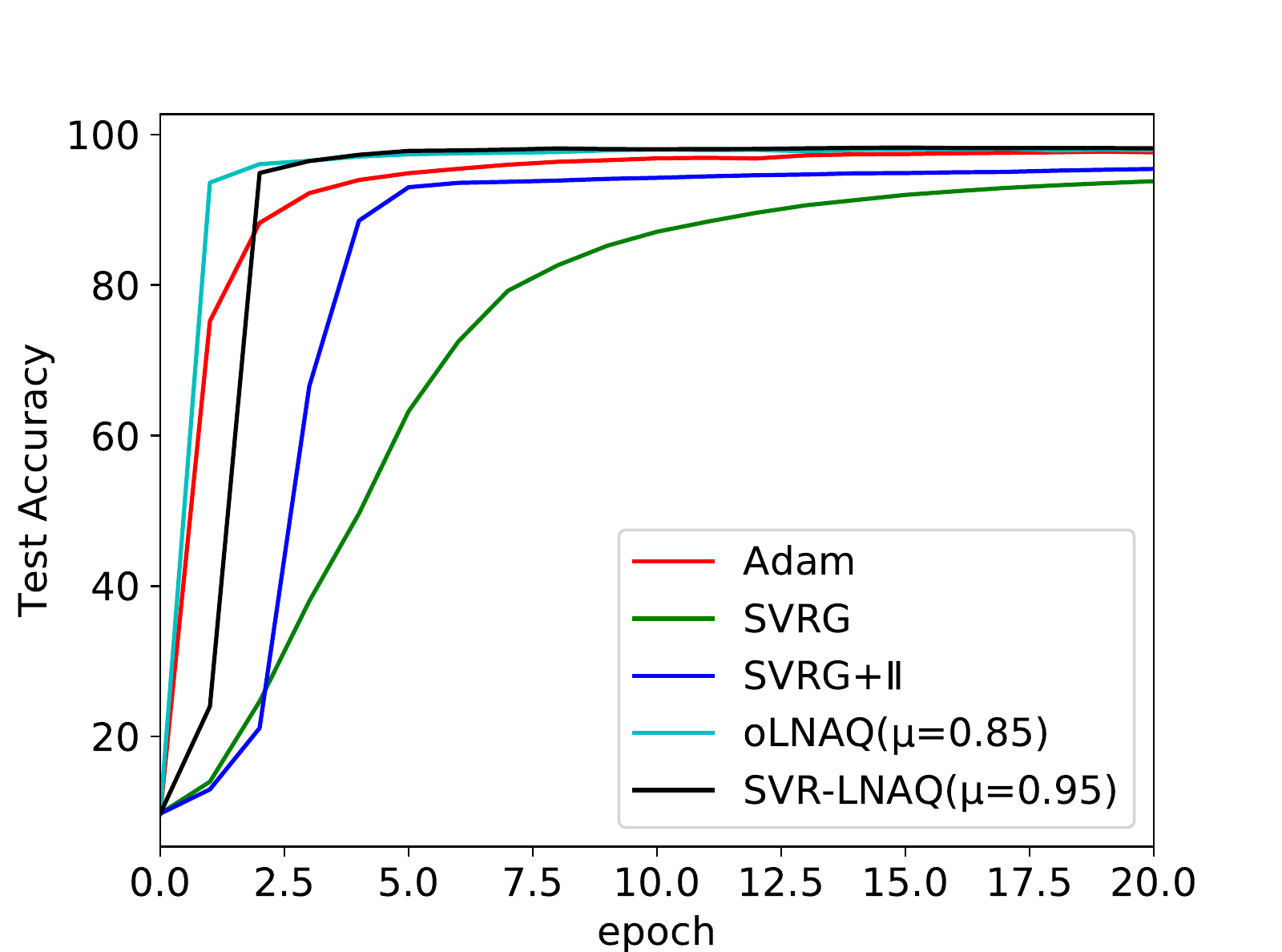}
\end{center}
\vspace{-5mm}
\caption{Test accuracy of MNIST for $b = 64$}
\label{fig:mnistTEA}
\vspace{-6mm}
\end{figure}

\begin{figure*}[ht]
\begin{minipage}[t]{0.49\textwidth}
\includegraphics[width=8cm]{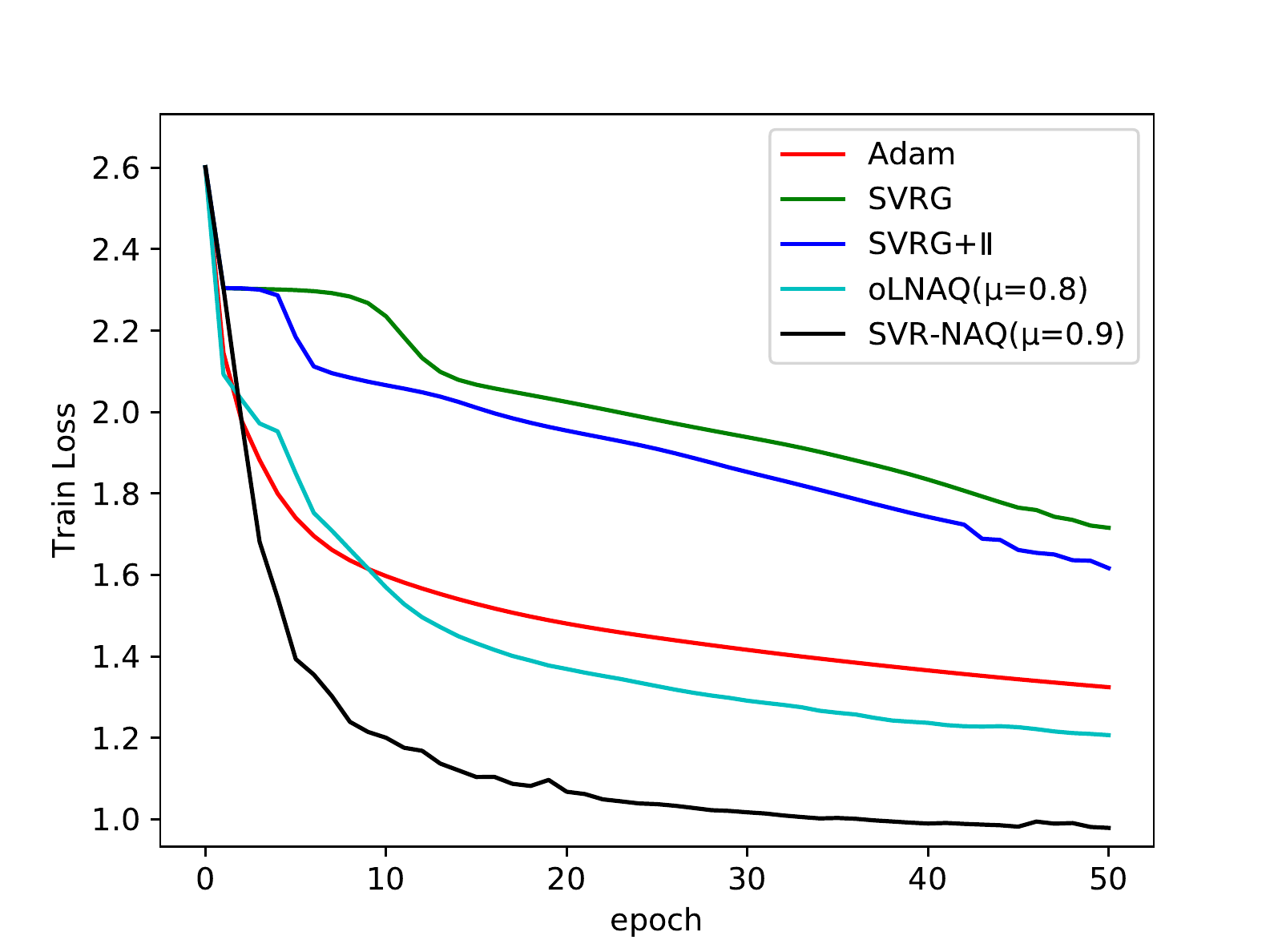}
\vspace{-3mm}
\caption{Train loss of CIFAR-10 for $b = 128$.}
\label{fig:cifarTRL}
\end{minipage}
\begin{minipage}[t]{0.49\textwidth}
\includegraphics[width=8cm]{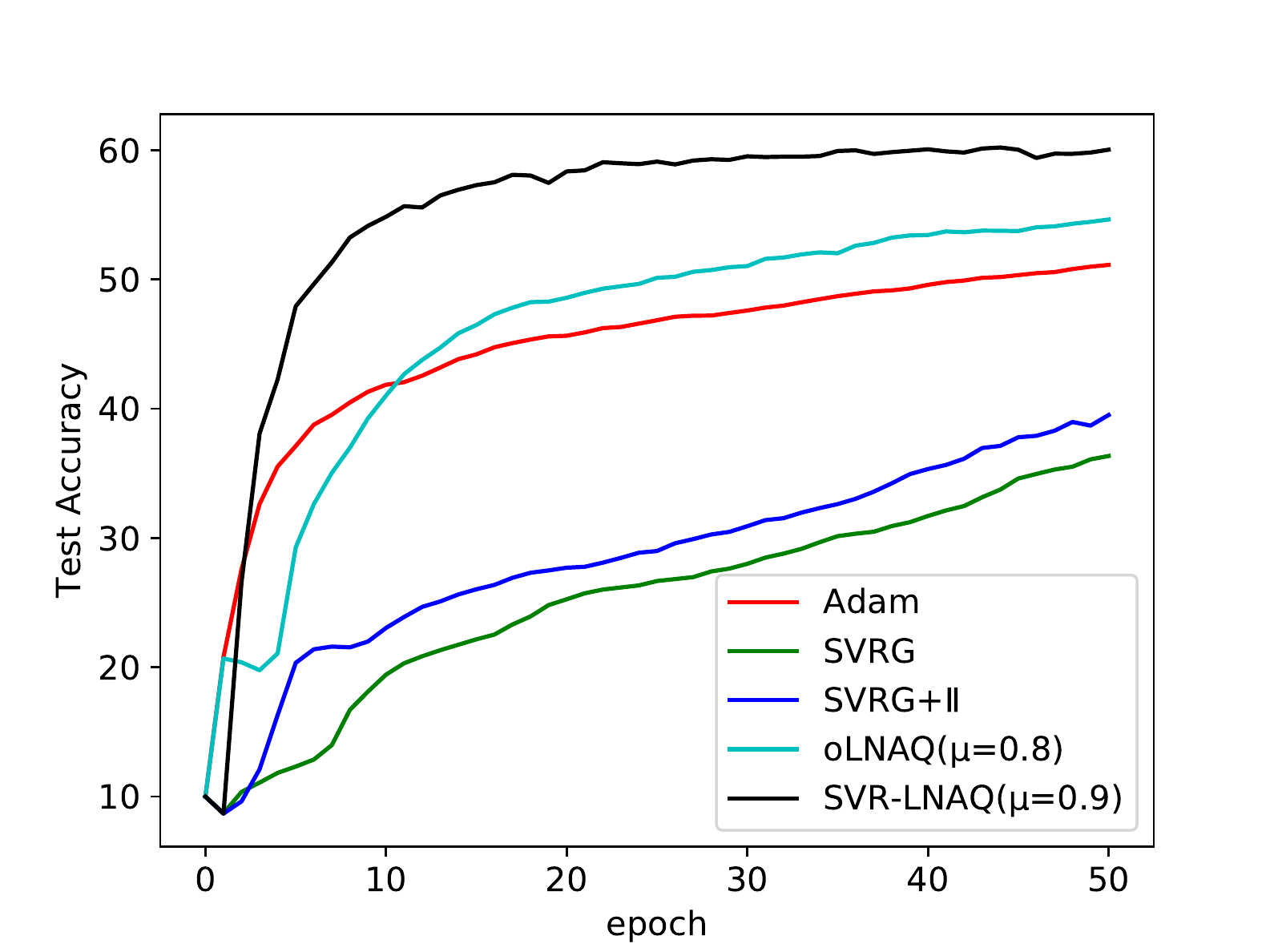}
\vspace{-2mm}
\caption{Test accuracy of CIFAR-10 for $b = 128$.}
\label{fig:cifarTEA}
\end{minipage}
\vspace{-4mm}
\end{figure*}

\subsubsection{Results on MNIST dataset}
The MNIST dataset is a standard benchmark image classification problem, where each sample comprises of a 28x28 pixel image of a handwritten digit. The CNN architecture comprises of two convolution layer of 3 and 5 3x3 filters respectively,each followed by 2x2 max pooling layer with stride 2. The convolution layers are followed by a fully connected layer with 100 hidden neurons. The batch size is chosen to be $b=64$ and momentum $\mu=0.85$ and $\mu=0.95$ for oLNAQ and SVR-LNAQ respectively. The simulations were carried out for a maximum of 25 epochs. The train loss and test accuracy are shown in Fig. \ref{fig:mnistTRL} and Fig. \ref{fig:mnistTEA} respectively. It can be observed that the proposed SVR-LNAQ is much faster in attaining a low train error and high test accuracy compared to SVRG, Adam, and SVRG-II.  In this problem again, it can be observed that though oLNAQ is faster in the initial epochs, the proposed SVR-LNAQ gives much lower error as the epochs increase. 

\subsubsection{Results on CIFAR-10 dataset}
Further, the performance of the proposed method is evaluated on a bigger problem. The CIFAR-10 is a 32x32 pixel 3 channel image classification problem. The train loss and test accuracy are shown in Fig. \ref{fig:cifarTRL} and Fig. \ref{fig:cifarTEA} respectively over 50 epochs. The batch size is chosen to be $b=128$ and momentum 0.8 and 0.9 for oLNAQ and SVR-LNAQ respectively. It was observed that for a bigger problem, the proposed SVR-LNAQ clearly outperforms SVRG, Adam, oLNAQ and SVRG-II. It can be seen to give much lower train errors and test accuracy much faster compared to the other algorithms, hence validating its efficiency.

\section{Conclusion}
A stochastic variance reduced Nesterov's Accelerated Quasi-Newton method  $-$ SVR-(L)NAQ is proposed both in its limited and full memory forms. The performance of the proposed method has been evaluated on regression and classification problems with both multi-layer neural network and convolution neural network respectively. The results show improved performance compared to SVRG, Adam, o(L)NAQ and SVRG-II methods. The introduction of variance reduction along with Nesterov's acceleration is shown to further accelerate convergence even for smaller batch sizes. 

In this paper, a simple adaptive step size is used. However, choice of the momentum term is another hyperparameter which could be determined using adaptive momentum selection schemes as in \cite{adaNAQ}. Also, training on larger neural networks and bigger problems can further test the limits of the proposed algorithm.  As future works, further possibilities of acceleration such as in  \cite{nitanda2014stochastic} and the convergence properties will be analyzed to provide proof of concept.   

\bibliography{references}{}
\bibliographystyle{ieeetr}

\end{document}